%% file: main.tex
\documentclass[conference]{IEEEtran}
\IEEEoverridecommandlockouts

\usepackage{cite}
\usepackage{amsmath,amssymb}
\usepackage{graphicx}
\usepackage{xcolor}
\usepackage{booktabs}
\usepackage{tabularx}
\usepackage{threeparttable}
\usepackage{hyperref}
\definecolor{mirablue}{RGB}{0,101,205}
\hypersetup{colorlinks=true, bookmarks=true, citecolor=mirablue}
\usepackage{pifont}
\usepackage{makecell}
\definecolor{okgreen}{RGB}{0,140,0}
\definecolor{nored}{RGB}{200,0,0}
\newcommand{\yes}{\textcolor{okgreen}{\checkmark}}
\newcommand{\no}{\textcolor{nored}{\ding{55}}}

\begin{document}

\title{MIRA: A Modular Open-Source Micro-UAV for Indoor Research\\
\thanks{Design files and software are available at \href{https://github.com/Lucas-Kido/mira.git}{\nolinkurl{github.com/Lucas-Kido/mira}}}
\thanks{\textsuperscript{*}These authors contributed equally to this work.}
\thanks{Correspondence to \href{mailto:kido.lucas0413@usp.br}{\fontfamily{pcr}\selectfont kido.lucas0413@usp.br}}
}

\author{\parbox{0.98\textwidth}{\centering\setlength{\tabcolsep}{3pt}%
    \begin{tabular}{*{4}{>{\centering\arraybackslash}p{1.63in}}}
    Lucas K. de Oliveira\textsuperscript{*} &
    Felipe Tommaselli\textsuperscript{*} &
    João Aires Marsicano\textsuperscript{*} &
    Marco S. Tayar \\
    \textit{Univ. of São Paulo (USP)} &
    \textit{Univ. of São Paulo (USP)} &
    \textit{Univ. of São Paulo (USP)} &
    \textit{Univ. of São Paulo (USP)} \\
    São Carlos, Brazil & São Carlos, Brazil &
    São Carlos, Brazil & São Carlos, Brazil \\
    \end{tabular}\\[1.2ex]
    \begin{tabular}{*{4}{>{\centering\arraybackslash}p{1.63in}}}
    Pedro A. R. Saraiva & Gianluca Capezzuto & Ricardo V. Godoy & Marcelo Becker \\
    \textit{Univ. of São Paulo (USP)} &
    \textit{Univ. of São Paulo (USP)} &
    \textit{Univ. of São Paulo (USP)} &
    \textit{Univ. of São Paulo (USP)} \\
    São Carlos, Brazil & São Carlos, Brazil & São Carlos, Brazil & São Carlos, Brazil \\
    \end{tabular}}
}

\maketitle

\begin{abstract}
Indoor robotics research increasingly relies on micro-UAV platforms whose airframes, electronics, and control software are open to modification. Off-the-shelf platforms often lack the low-level access that such modifications require, while building a custom vehicle demands substantial engineering effort before the first flight test, leaving many laboratories working within constraints that limit the scope of their research. We present MIRA (Modular Indoor Research Architecture), an open-source micro-UAV for indoor research, built around a replicable 3D-printed PLA airframe with detachable arms and a containerized low-level software package that manages the companion-to-autopilot communication bridge via Micro XRCE-DDS. Designed as a white-box architecture, its core subsystems, including the companion computer, are individually replaceable without firmware refactoring, which supports local fabrication, crash-localized repair, and component substitution from existing laboratory inventory. We characterize MIRA through autonomous flights comprising takeoff, hovering, trajectory tracking, and landing within an optical motion-capture volume. The communication pipeline sustains a median companion-to-autopilot latency of 0.02 ms, with the external motion-capture link as the main remaining latency source, and time-domain analysis shows that structural vibration remains stable during dynamic maneuvers and below the reference level adopted in this work. All design files and software are released openly to support replication and extension by other laboratories.
\end{abstract}


\input{sections/01_introduction}
\input{sections/02_related_works}
\input{sections/03_system_design}
\input{sections/04_results}
\input{sections/05_discussion}
\input{sections/06_conclusion}

\section*{Acknowledgment}

This work was supported by the São Paulo Research Foundation (FAPESP), Grants \#2025/20858-7, \#2025/08520-0 and \#2025/22381-3; and by Petróleo Brasileiro S/A - Petrobras, using resources from the ANP R\&D clause, in partnership with the University of São Paulo (USP) and the Fundação de Apoio à Física e à Química (FAFQ), under Cooperation Agreements \#2023/00016-6 and \#2023/00013-7.

\bibliographystyle{IEEEtran}
\bibliography{lars_bib}

\end{document}

%% file: sections/01_introduction.tex
\section{Introduction} \label{section:introduction}
\begin{figure}[!t]
    \centering
    \includegraphics[width=1\linewidth]{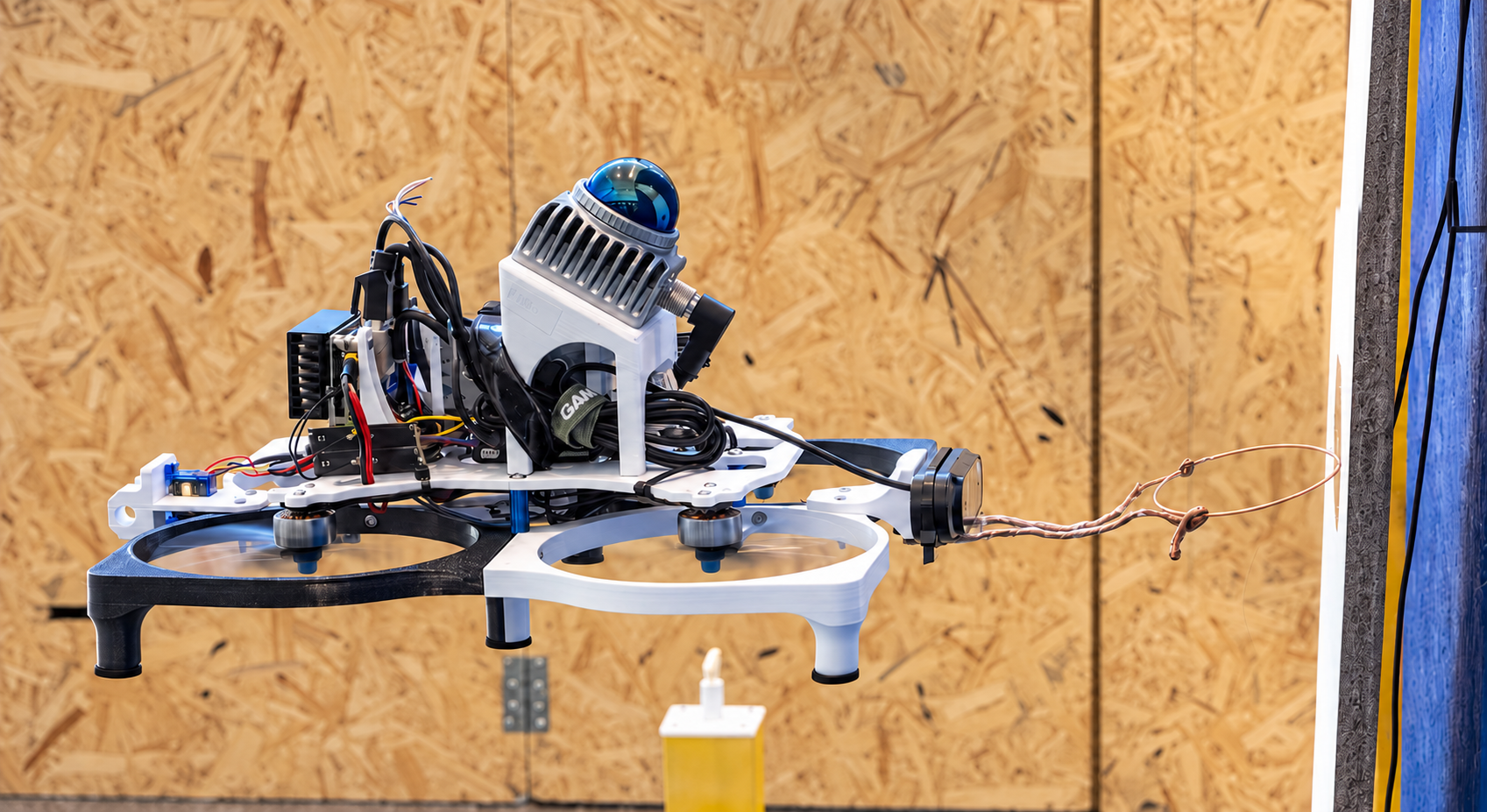}
    \caption{The assembled physical prototype of the MIRA quadrotor platform.}
    \label{fig:mira}
\end{figure}

Indoor aerial robotics research faces a persistent infrastructure bottleneck. Affordable commercial drones feature closed ecosystems and payload constraints that restrict custom hardware integration, while open academic platforms often mandate rigid hardware configurations that escalate replication costs and limit component substitution from existing lab inventory. This collective engineering and financial overhead scales poorly with fleet size, creating a barrier to entry for laboratories requiring replicable indoor testing baselines. Fig.~\ref{fig:mira} shows the platform proposed in this paper.

Although open-source flight stacks and ROS 2 middleware have lowered software barriers, the hardware modularity gap persists~\cite{meier2015px4, macenski2022ros2}. The Crazyflie 2.1 Brushless is a reference swarm vehicle, but its 40\,g payload limits onboard processing~\cite{preiss2017crazyswarm, bitcraze2025}. Conversely, larger academic platforms like Agilicious treat sensing, control, and compute as a jointly optimized, custom-PCB architecture~\cite{Foehn22science}. Similarly, advanced platforms such as OmniNxt~\cite{liu2024omninxt} and AgiPIX~\cite{agipix2026} remain bound to specific, co-developed hardware configurations or costly active sensing suites. A clear gap remains for a transparent, standard-component baseline that can be fabricated, repaired, and provisioned rapidly using common workshop tools.

This paper introduces a system that addresses this practical bottleneck by focusing entirely on providing accessible physical and digital infrastructure rather than optimizing for high agility or specific perception suites. The platform is simultaneously (i) optimized for indoor, GPS-denied environments, (ii) easily replicable on standard desktop FDM printers using PLA, (iii) modular at the component level to allow substitution of parts from existing lab inventory, and (iv) supported by a containerized low-level software stack.

Local manufacturability and hardware interchangeability operate as primary design constraints throughout the system. The airframe utilizes a detachable arm topology to localize crash damage, while the avionics interface relies on a robust, firmware-independent serial connection. On the software side, a streamlined, containerized ROS 2 package manages the low-level uXRCE-DDS communication bridge instead of a monolithic autonomy framework. This setup handles the essential task of ingesting external motion-capture pose streams and translating them into the autopilot's internal filter. The result is a clean white-box baseline that reduces engineering overhead, leaving onboard computational resources available for user-defined algorithms. 

The primary contributions of this work are:
\begin{itemize}
    \item \textbf{Open modular airframe.} A 3D-printed PLA quadrotor structure with a detachable interface, supporting local fabrication, rapid crash-localized repair, and fleet replication.
    \item \textbf{Hardware-agnostic communication bridge.} A containerized ROS 2 package over a serial Micro XRCE-DDS link that decouples low-level flight execution from high-level compute, allowing hardware substitution without firmware refactoring.
    \item \textbf{Streamlined state estimation pipeline.} An out-of-the-box software architecture deployed on a Jetson Orin NX, optimized to ingest real-time external pose streams and map them transparently to the PX4 EKF2 estimator.
    \item \textbf{Targeted physical validation.} Experimental flight characterization using autonomous trajectory tracking within an optical motion-capture environment, evaluating communication pipeline latency and structural vibration through time-domain metrics.
\end{itemize}

The rest of the paper is organized as follows: Section~\ref{sec:rw} presents an overview of the related works, Section~\ref{section:system_design} describes the platform's mechanical, electrical, and software design, Section~\ref{section:results} reports the experimental pipeline latency and structural results, Section~\ref{section:discussion} discusses the engineering trade-offs, and Section~\ref{section:conclusion} concludes the paper.

%% file: sections/02_related_works.tex
\section{Related Works}\label{sec:rw}
Open-source flight stacks~\cite{px4, meier2015px4} and robotics middleware~\cite{macenski2022ros2} have lowered the barrier to building research UAVs, yet scaling a fleet reveals a persistent issue: platforms optimized for single objectives are difficult to replicate, repair, or adapt. At the commercial end, the Crazyflie 2.1 Brushless offers open firmware, but its 40\,g payload limits onboard processing, forcing reliance on external infrastructure~\cite{preiss2017crazyswarm, bitcraze2025}. Conversely, commercial platforms like the DJI M400~\cite{dji_matrice400_specs} and Flybotix ASIO X~\cite{flybotix_asio_x} provide greater thrust and payload capacities, and the DJI Tello EDU~\cite{airswarm2025} offers programmable access, but all impose closed, proprietary firmware, thereby confining research to software layers operating above the flight controller ecosystem.

Academic platforms recover firmware transparency but introduce rigid hardware constraints. For instance, the FLA-Quad~\cite{FLA_QUAD} and MRS UAV~\cite{MRS} offer robust baselines but rely on specific component configurations that can restrict hardware substitution. Agilicious~\cite{Foehn22science} provides an open baseline for agile flight but treats sensing, control, and compute as a jointly optimized, single custom PCB, rendering component substitution difficult. OmniNxt~\cite{liu2024omninxt} enables omnidirectional perception, yet its companion computer and firmware form a fixed, co-developed configuration. AgiPIX~\cite{agipix2026} improves software reproducibility via containerized ROS 2 stacks but remains bound to hardware-synchronized active sensing and specific GPU modules. Consequently, while various open-source state estimation pipelines are accessible, their adoption on custom platforms is constrained by the lack of an adaptable low-level communication framework that can ingest and map external pose data to the autopilot.

Thus, existing systems optimize openness along a single axis, such as firmware, perception, or software, while hardware-level interchangeability and local manufacturability remain secondary. This work addresses this gap by proposing an indoor-oriented platform that is simultaneously: (i) locally manufacturable via common desktop FDM printers, (ii) modular at the component level to allow inventory-driven substitution of flight controllers and companion computers, and (iii) supported by a containerized low-level bridge designed for external state estimation ingest.

\begin{figure*}[!t]
    \centering
    \includegraphics[width=0.6\linewidth]{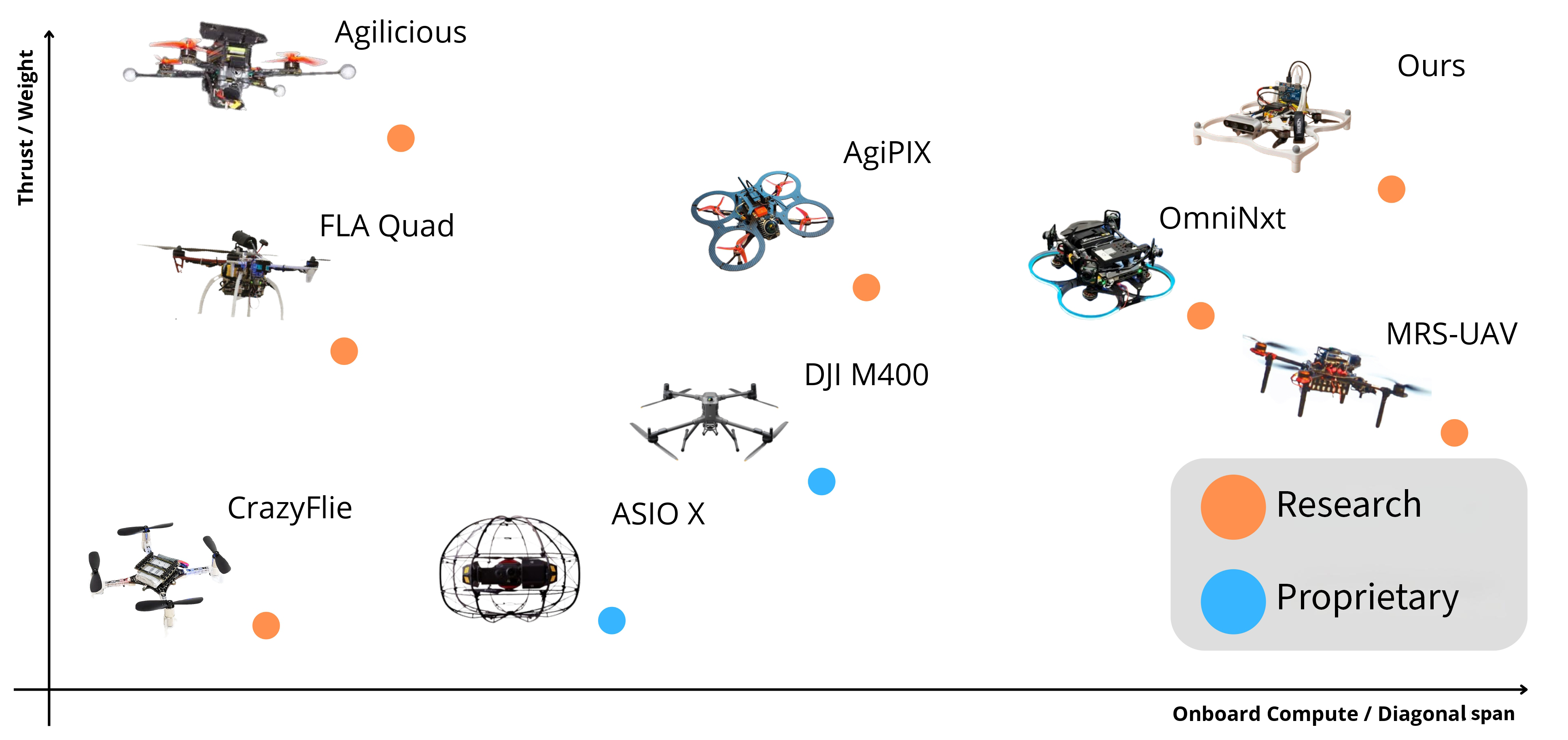}
    \caption{Schematic positioning of the surveyed platforms by thrust-to-weight ratio (vertical axis) and onboard compute relative to diagonal span (horizontal axis). Orange markers denote research platforms and blue markers denote proprietary platforms; the underlying values are listed in Table~\ref{tab:drone_comp}.}
    \label{fig:landscape}
\end{figure*}

To illustrate the platform's focus, a feature matrix comparing these existing systems with the proposed MIRA architecture is summarized in Table~\ref{tab:drone_comp}, showing how MIRA bridges the gap between hardware openness and local manufacturability.
Fig.~\ref{fig:landscape} complements Table~\ref{tab:drone_comp} by placing the surveyed platforms in the thrust-to-weight versus compute-and-size plane.

\begin{table*}[!t]
\centering
\caption{Comparison of Consumer and Research Platforms. Additional Metrics Include Openness, ROS 2 Support, Simulation Capabilities (Double Ticks Denote Photorealism), CPU Performance, and GPU Availability.}
\label{tab:drone_comp}
\begin{threeparttable}
\footnotesize
\setlength{\tabcolsep}{3.2pt}
\begin{tabular}{@{}lcccccccccc c@{}}
\toprule
\textbf{framework} &
\textbf{open-source} &
\textbf{ROS 2} &
\textbf{simulation} &
\textbf{\makecell{low-level \\ controller}} &
\textbf{\makecell{CPU mark \\ (higher is \\ better)}} &
\textbf{GPU} &
\textbf{\makecell{3D \\ LIDAR}} &
\textbf{DDS} &
\textbf{\makecell{collision \\ guard}} &
\textbf{\makecell{thrust / \\ weight}} &
\textbf{\makecell{diagonal \\ span (mm)}} \\
\midrule
DJI M400 \cite{dji_matrice400_specs} & - & \yes & \no & proprietary & - & \yes & \yes & \no & \no & - & $\approx 1240$ \\
Flybotix ASIO X \cite{flybotix_asio_x} & - & \no & \no & proprietary & - & \no & \yes & \no & \no & - & - \\
Crazyflie \cite{bitcraze2025} & SW and HW & \yes & \yes & custom & - & \no & \no & \no & \no & $\approx 2.26$ & $\approx 92$ \\
FLA-Quad \cite{FLA_QUAD} & SW and HW & \no & \yes & PX4 & 3,383 & \no & \yes & \no & \no & $\approx 2.38$ & - \\
MRS UAV \cite{MRS} & SW and HW & \yes & \yes & PX4 & 9,264 & \no & \yes & \no & \no & $\approx 2.5$ & $\approx 792$ \\
Agilicious \cite{Foehn22science}  & SW and HW\tnote{a} & \no & \yes\,\yes & custom & 1,343 & \yes & \no & \no & \no & $\approx 5$ & $\approx 382$ \\
OmniNxt \cite{liu2024omninxt} & SW and HW & \yes & \no & PX4 & 2,418 & \yes & \no & \no & \yes & $\approx 4.24$ & $\approx 250$ \\
AgiPIX \cite{agipix2026} & SW and HW & \yes & \yes\,\yes & PX4 & 2,418 & \yes & \yes & \yes & \yes & $\approx 3.5$ & $\approx 495$ \\
\midrule
Ours & SW and HW & \yes & \yes & PX4 & 2,418\tnote{b} & \yes & \textcolor{blue}{$\square$}\tnote{c} & \yes & \yes & $\approx 5$ & $\approx 229$ \\
\bottomrule
\end{tabular}
\begin{tablenotes}\footnotesize
\item[a] Conditionally open-source.
\item[b] Jetson Orin NX baseline; the companion computer is replaceable (Table~\ref{table:bom}).
\item[c] Interface supports 3D LiDAR; not implemented in this baseline.
\end{tablenotes}
\end{threeparttable}
\end{table*}

%% file: sections/03_system_design.tex
\section{System Design}
\label{section:system_design}

\subsection{Architecture Overview}
\begin{figure*}[!t]
    \centering
    \includegraphics[width=0.80\linewidth]{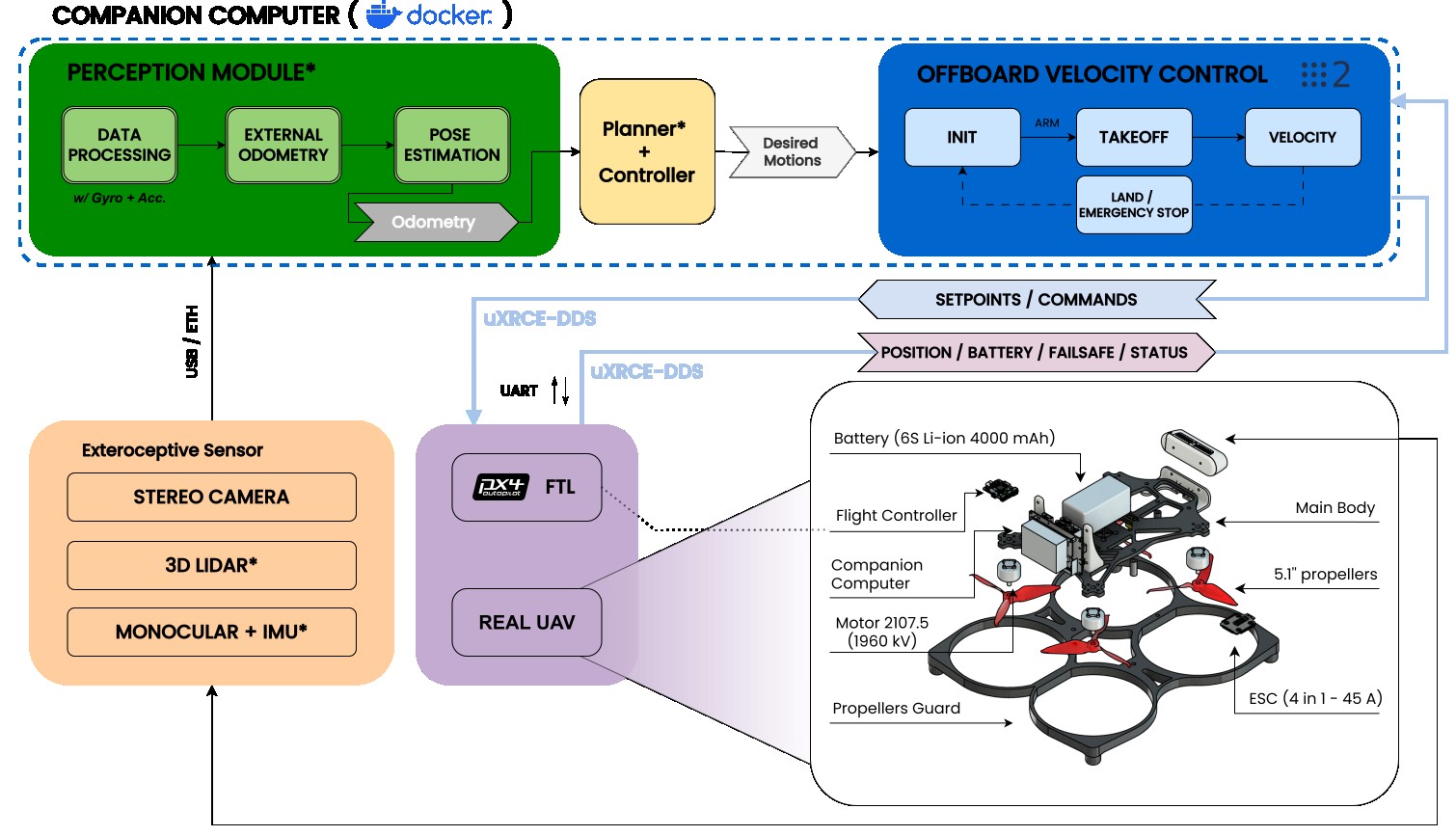}
    \caption{Breakdown of the UAV system, illustrating the functional layers from the containerized software communication bridge to the exploded hardware assembly. Modules marked with an asterisk (*) and the listed exteroceptive sensors are not implemented in this paper and illustrate possible platform extensions.}
    \label{fig:drone_diagram}
\end{figure*}

The platform follows a modular three-layer architecture designed for hardware agnosticism, allowing individual layers to be maintained or entirely substituted based on user requirements (Fig.~\ref{fig:drone_diagram}). The real-time control layer runs the PX4 autopilot on a flight controller to handle low-level control loops and sensor fusion, while the perception and offboard layer runs on a companion computer that hosts ROS 2 for high-level processing. Connecting these layers, the communication bridge uses the Micro XRCE-DDS protocol over a serial link, completely decoupling flight execution from compute processes and enabling hardware swaps without refactoring the core software stack.

\subsection{Mechanical Design}
The 3D-printed airframe consists of a central avionics deck and four detachable arms. PLA was selected to enable on-demand replication using standard desktop FDM printers. While this material choice inherently introduces trade-offs in structural stiffness, weight, and long-term durability, the modular interface acts as an effective crash-mitigation mechanism, enabling damaged arms to be rapidly replaced. To compensate for material flexibility, structural components are printed with a 15\,\% honeycomb infill and four perimeter walls. This configuration is intended to place structural resonance frequencies outside the operational bandwidth of the attitude control loops; a frequency-domain verification is left for future work (Section~\ref{section:discussion}), and Section~\ref{subsec:vibration} reports the time-domain vibration levels measured in flight.

\subsection{Electronics and Power}

The electronics layer is structured around strict component interchangeability. The Pixhawk 6C Mini was selected for its compact form factor and ecosystem support, although its internal electronics are not entirely transparent; researchers requiring fully open hardware can substitute any board that provides the necessary serial interfaces. Power distribution relies entirely on commercial off-the-shelf (COTS) parts: a standard 4-in-1 ESC and direct battery connections with integrated BECs, which eliminates custom wiring harnesses. Table~\ref{table:bom} lists the baseline used in flight together with unit costs.

\begin{table}[!b]
\centering
\caption{Bill of Materials with Unit Costs and Companion Computer Options (US\$, Vendor Listings Accessed in October 2026).}
\label{table:bom}
\begin{threeparttable}
\footnotesize
\setlength{\tabcolsep}{3pt}
\begin{tabularx}{\columnwidth}{@{}l X r r@{}}
\toprule
\textbf{Component} & \textbf{Product and specification} & \textbf{Cost} & \textbf{Total} \\
\midrule
Frame & PLA (FDM), 15\% infill & $\approx$5\tnote{a} & \\
Motor ($\times$4) & GEPRC SpeedX2 2107.5, 1960\,KV \cite{geprc_speedx2} & 96\tnote{b} & \\
Propeller ($\times$4) & T-Motor T5146 \cite{tmotor_t5146} & 6\tnote{c} & \\
Battery & iFlight Fullsend Li-ion, 6S, 4000\,mAh \cite{iflight_fullsend} & 50 & \\
ESC & Hobbywing XRotor 45A 4-in-1 \cite{hobbywing_xrotor} & 115 & \\
Flight controller & Pixhawk 6C Mini, STM32H743 \cite{holybro_6cmini} & 179 & \\
\midrule
\multicolumn{3}{@{}l}{\textit{Flight core subtotal}} & \textit{451} \\
\midrule
\multicolumn{4}{@{}l}{\textit{Companion computer (select one)\tnote{d}}} \\
Jetson Orin NX & 16\,GB module (flown) \cite{nvidia_jetson_faq} & 999 & 1,450 \\
Jetson Orin Nano & 8\,GB module (not flown) \cite{nvidia_jetson_faq} & 399 & 850 \\
Raspberry Pi 5 & 8\,GB (not flown) \cite{pishop_rpi5} & 175 & 626 \\
Raspberry Pi 5 & 4\,GB (not flown) \cite{pishop_rpi5} & 110 & 561 \\
\bottomrule
\end{tabularx}
\begin{tablenotes}\footnotesize
\item[a] Filament only (estimate).
\item[b] Listing price (US\$23.99 each) shared by the 1960 and 2450\,KV versions.
\item[c] Not listed in T-Motor's store; retailer price.
\item[d] Total = flight core + companion. Jetson: NVIDIA 1k-unit prices (single-unit prices are higher); carrier board, storage, and cooling are not included.
\end{tablenotes}
\end{threeparttable}
\end{table}

\subsection{Software and Communication}
\label{subsec:software}
The high-level software stack is deployment-agnostic, running containerized environments via Docker to guarantee reproducibility. ROS 2 and the Micro XRCE-DDS agent run inside these containers, which isolates their dependencies from the companion computer’s operating system and simplifies replication in other laboratories.

\textbf{State Estimation Interface.}
The software architecture exposes a generalized interface for external state estimation. Any tracking system, such as visual-inertial odometry (VIO) or motion capture, can provide spatial data by publishing standard ROS 2 odometry topics, which are mapped to the standard \texttt{VehicleOdometry} uORB topics. MIRA currently relies on PX4; other firmwares, such as ArduPilot, may expose message-mapping issues within the uXRCE-DDS bridge.

\textbf{Offboard Control Node.}
Command arbitration is managed by a dedicated and highly customizable ROS 2 C++ node. Depending on the specific research needs, this node can be easily configured to control the drone through position, velocity, or direct attitude setpoints. It dynamically rotates body-frame velocity commands to the absolute NED frame before transmission. A latency watchdog and an emergency stop service lock subsequent inputs and command a safe landing routine via MAVLink. The serial Micro XRCE-DDS link is hardware-agnostic but adds delay that may limit highly dynamic tasks.

\subsection{Compute Selection and Cost}
\label{subsec:cost}
MIRA releases a mechanical structure and a software bridge, not a fixed compute configuration: the bridge only requires a companion computer that runs Docker with ROS 2 and exposes a serial port. We used the Jetson Orin NX 16\,GB because our target application runs neural-network detectors (YOLO-family) on camera streams, which requires GPU acceleration; the flights reported here use motion-capture pose only. Pose-based flight and offboard control can use a single-board computer such as a Raspberry Pi 5, which we did not flight-validate (Table~\ref{table:bom}).

The compute unit dominates the cost: with a flight core of about US\$451, the Orin NX module is 69\,\% of the baseline and a Raspberry Pi 5 (8\,GB) is 28\,\%. Prices are also volatile: the Orin NX list price rose from US\$599 to US\$999 in July 2026 \cite{nvidia_jetson_faq}. For reference, the Crazyflie 2.1 Brushless costs US\$480 without a Linux-class companion \cite{bitcraze_store}, and the Holybro X500 V2 PX4 kit US\$609 without battery and companion \cite{holybro_x500}. MIRA with a Raspberry Pi 5 thus falls in the price range of such kits, while the Orin NX configuration costs about 2.3 times as much and is justified only by onboard GPU inference. For a fleet of ten vehicles, these two configurations amount to about US\$14,500 and US\$6,300, respectively, so GPU companions are best reserved for the vehicles that run onboard inference. MIRA's advantage lies less in unit price than in local fabrication, crash-localized repair, and component substitution.

%% file: sections/04_results.tex
\section{Results} \label{section:results}

\subsection{Experimental Setup}
Experiments were conducted in an indoor GPS-denied environment using an OptiTrack PrimeX motion-capture system, providing millimeter-level ground-truth and real-time pose estimates. A 2.4\,GHz remote controller served only as a safety kill switch; all flight phases (takeoff, hovering, tracking, and landing) were executed autonomously by the companion computer. Communication timing statistics are reported in Table~\ref{tab:jitter}.

\subsection{Autonomous Flight Performance and Communication Latency}
\label{subsec:latency}

\begin{figure}[!t]
    \centering
    \includegraphics[width=\columnwidth]{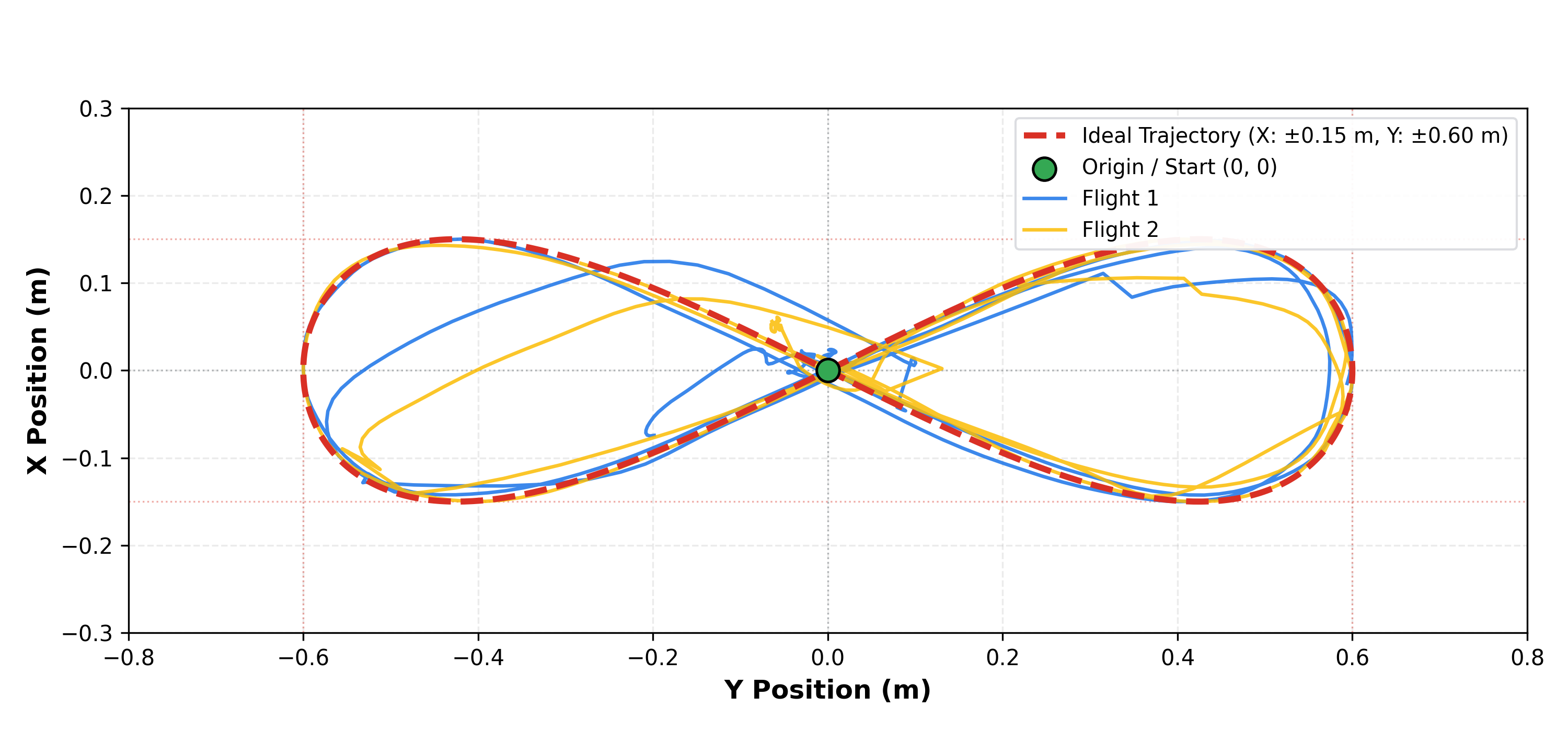}
    \caption{Top-down 2D spatial plot comparing the ideal figure-8 trajectory reference against the actual autonomous flight executions.}
    \label{fig:trajectory_tracking}
\end{figure}

To evaluate closed-loop control and the ROS 2 to PX4 bridge, the platform executed a dynamic figure-8 trajectory ($0.30 \text{ m} \times 1.20 \text{ m}$). As illustrated in Fig.~\ref{fig:trajectory_tracking}, both flights follow the reference trajectory closely. Minor deviations at the cross-section and sharp curves are expected characteristics of standard cascaded PID controllers tracking aggressive setpoints without predictive feedforward terms. Furthermore, a transient position error spike at the end corresponds exclusively to landing and motor disarming, not affecting in-flight tracking accuracy.

Regarding the data pipeline, the flight log was analyzed to evaluate communication timing (Table~\ref{tab:jitter}). The motion-capture input reached the FMU at 173.6\,Hz, and EKF2 published \texttt{vehicle\_odometry} at 93.2\,Hz. The pipeline sustained a median companion-to-autopilot latency of 0.02\,ms. The P95 (5.34\,ms) and P99 (39.12\,ms) latencies represent occasional uXRCE-DDS transport overhead. A critical bottleneck appeared externally: \texttt{/drone/pose} exhibited a median latency of 554\,ms, exposing a clock misalignment between the NatNet/ROS bridge and the host.

\begin{table}[!t]
\centering
\caption{ROS 2 Topic Statistics from the Autonomous Flight Log}
\label{tab:jitter}
\begin{threeparttable}
\footnotesize
\setlength{\tabcolsep}{3pt}
\begin{tabular}{@{}l c c c@{}}
\toprule
\textbf{Topic} & \textbf{Freq. (Hz)} & \textbf{CV (\%)\tnote{a}} & \textbf{Max gap (ms)} \\
\midrule
OptiTrack Odometry (in) & 173.6 & 142.0 &  175.9 \\
Sensor Combined (IMU)   & 146.4 & 609.6 & 2060.4 \\
Vehicle Attitude        & 146.2 & 610.9 & 2060.3 \\
FMU Odometry (out)      &  93.2 & 488.8 & 2053.0 \\
Vehicle Local Position  &  93.2 & 488.2 & 2053.3 \\
Drone Pose              & 176.5 & 252.1 & \textbf{2984.6} \\
\bottomrule
\end{tabular}
\begin{tablenotes}\footnotesize
\item[a] CV (\%) denotes temporal jitter. The shared $\approx$2060\,ms gap across four FMU topics identifies a single EKF2 reset event.
\end{tablenotes}
\end{threeparttable}
\end{table}

\subsection{Structural Vibration Analysis}
\label{subsec:vibration}
\begin{figure*}[!t]
    \centering
    \includegraphics[width=0.75\linewidth]{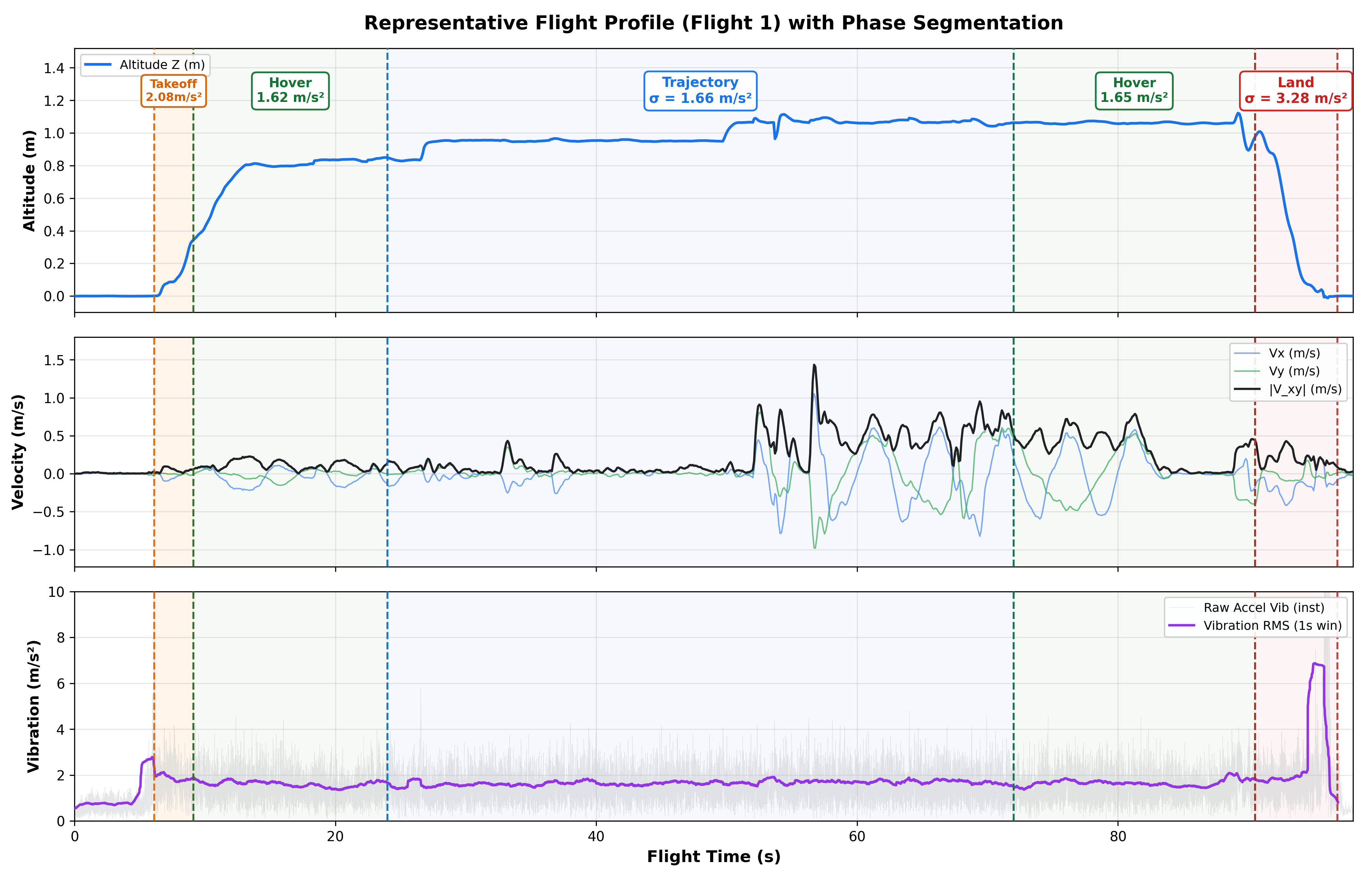}
    \caption{Representative autonomous flight segmented into takeoff, hover, trajectory tracking, and landing: altitude (top), velocity (middle), and accelerometer vibration (bottom).}
    \label{fig:flight_profile}
\end{figure*}

Actuator outputs display symmetry across all channels during steady-state conditions, tracking a baseline hover throttle of $\approx 20\,\%$. To evaluate the structural integrity across different dynamic regimes, a time-domain analysis of the segmented flight profile was conducted (Fig.~\ref{fig:flight_profile}). The standard deviation of the accelerometer data ($\sigma$) was used as the primary metric, in line with standard autopilot evaluation criteria.

During the takeoff phase, vibration increased due to ground effect turbulence. During the trajectory tracking phase, where velocity changes, the vibration levels remained stable ($\sigma = 1.66 \text{ m/s}^2$), matching the baseline hover performance ($\sigma \approx 1.62 \text{ m/s}^2$). The highest recorded vibration ($\sigma = 3.28 \text{ m/s}^2$) occurred upon touchdown. During free flight, these values stay below the reference level of $2.0 \text{ m/s}^2$ adopted in this work for the standard deviation of the accelerometer signal; PX4 itself provides only qualitative guidance on acceptable vibration levels \cite{px4_flight_review}. In the representative flight analyzed here, the 15\% infill PLA airframe therefore supports state estimation without additional external mechanical dampeners, with the flight controller's built-in vibration isolation left in place.

%% file: sections/05_discussion.tex
\section{Discussion} \label{section:discussion}

The platform prioritizes local manufacturability and modular component reuse over specialized sensing or high agility, giving a transparent, easily repairable baseline vehicle. Although the serial uXRCE-DDS bridge adds transport delay compared to PCIe or USB interconnects, the recorded latency remains acceptable for stable state fusion and standard trajectory tracking.

\textbf{Limitations.} Evaluation is currently confined to controlled indoor environments, and only the Jetson Orin NX companion was flight-validated. The experiments revealed an external bottleneck: the network-level motion-capture bridge introduced noticeable delay and clock misalignment, dominating the latency budget relative to the drone's internal pipelines. Onboard exteroceptive sensors, such as VIO cameras or 3D LiDAR, add payload weight and must be evaluated against the thrust-to-weight ratio and control bandwidth.

\textbf{Future work.} Future steps include resolving the external clock alignment, for example through time synchronization between the motion-capture host and the companion computer, and evaluating onboard state estimation through a VIO pipeline. Onboard estimation removes the dependence on external infrastructure, but VIO adds CPU/GPU load and latency, so a net latency reduction is not guaranteed and will be quantified. We also plan a frequency-domain analysis of raw IMU data (e.g., power spectral density) to verify the resonance behavior of the airframe, a comparison of vibration across infill percentages, flight validation of a Raspberry Pi 5 companion, and multi-agent fleets.

%% file: sections/06_conclusion.tex
\section{Conclusion} \label{section:conclusion}
We presented MIRA, an open-source modular micro-UAV architecture for GPS-denied indoor research. By combining a locally manufacturable PLA airframe with a hardware-agnostic PX4, ROS 2, and uXRCE-DDS low-level communication bridge, the platform offers a transparent, white-box baseline that is straightforward to repair, replicate, and extend, with a flight core of about US\$450 and a companion computer selected per application. Autonomous flights confirm closed-loop tracking, vibration levels below our reference with 15\% infill, and a high-rate data pipeline whose main latency bottleneck is the external motion-capture bridge. All mechanical design files (3D models), software development configurations (Docker containers and ROS nodes), and documentation are released openly to support reproducible research in indoor aerial robotics.